\documentclass[conference]{IEEEtran}
\IEEEoverridecommandlockouts

\usepackage{cite}
\usepackage{amsmath,amssymb,amsfonts}
\usepackage{graphicx}
\usepackage{textcomp}
\usepackage{xcolor}

\def\BibTeX{{\rm B\kern-.05em{\sc i\kern-.025em b}\kern-.08em
    T\kern-.1667em\lower.7ex\hbox{E}\kern-.125emX}}

\usepackage{algorithm}
\usepackage{algpseudocode} 
\usepackage{multirow,tabularx}

\begin{document}

\title{Learning Rate Optimization for Deep Neural Networks Using Lipschitz Bandits
}

\author{
    \IEEEauthorblockN{%
        Padma Priyanka, %
        $\, $%
        Sheetal Kalyani, Avhishek Chatterjee}
    \IEEEauthorblockA{\textit{Department of Electrical Engineering,} \\
        \textit{Indian Institute of Technology Madras,}
        \textit{
            Chennai - 600036, 
            India.
        }
        \\
        e-mail: \{ee18d035@smail, skalyani@ee, avhishek@ee\}.iitm.ac.in}
}

\maketitle

\begin{abstract}

Learning rate is a crucial parameter 
 in training of neural networks. A properly tuned learning rate leads to faster training and higher test accuracy. In this paper, we propose a Lipschitz bandit-driven approach for tuning the learning rate of neural networks.
The proposed approach is compared with the popular HyperOpt technique used extensively for hyperparameter optimization and the recently developed bandit-based algorithm BLiE. The results for multiple neural network architectures indicate that our method finds a better learning rate using a) fewer evaluations and b) lesser number of epochs per evaluation,  when compared to both HyperOpt and BLiE. Thus, the proposed approach enables more efficient training of neural networks, leading to lower training time and lesser computational cost.
\end{abstract}

\begin{IEEEkeywords}
Neural network training; learning rate; 
\end{IEEEkeywords}

\section{Introduction}
Deep neural networks have many applications in computer vision \cite{liu2017survey}, communications \cite{mao2018deep}, etc. Parameters of a neural network model are learned during training. A neural network's hyperparameters, such as width, depth, and learning rate, that are chosen for the training process significantly influence the training procedure and the performance of the learned model. 
\par Sequential model-based optimization (SMBO) was introduced to perform hyperparameter tuning since conventional grid and random search are inefficient. In SMBO, hyperparameters are updated based on the past observations \cite{snoek2012practical}. Some of the well-known SMBO methods that use Bayesian approach are Gaussian processes \cite{rasmussen2003gaussian}, random forests \cite{hutter2011sequential}, tree-structured Parzen estimator (TPE) \cite{bergstra2011algorithms} and HyperOpt \cite{bergstra2015hyperopt}. HyperOpt harnesses both random search and TPE as its subroutines and is the current state of the art. 
There is another class of methods where hyperparameter optimization is modelled as a multi-arm bandit problem.  Hyperband uses a successive halving \cite{jamieson2016non} method to discard bad configurations \cite{li2018hyperband} and  Bayesian Optimization and Hyperband (BOHB)  amalgamates Hyperband and Bayesian optimization  \cite{falkner2018bohb}. 
\par 
In many applications, the network architecture, i.e., depth, width and connections, is fixed. Learning rate is the central hyperparameter in those scenarios as it directly influences the step size of the optimization algorithm (e.g., gradient descent and its variants)  used in training and thus influencing convergence and test accuracy \cite{senior2013empirical}. As observed before, the optimal learning rate is directly related to the Lipschitz constant of the gradient of the training objective  \cite{nesterov2004introductory}. 
\par An analytical approach for determining the gradient Lipschitz constant of a shallow feed-forward neural
network was proposed in \cite{Tholeti_2020}. They derived the gradient Lipschitz constant for shallow networks ($1$ and $2$ layers). However, scaling that approach to networks with more than two layers is fairly difficult. A natural extension of the above approach to networks with multiple layers would involve approximating the gradient Lipschitz constant and then choosing the learning rate accordingly.  
\par This work, though inspired by the Lipschitz properties inherent in the training procedure, takes a  different approach. It is generally observed that the average training loss and the average test accuracy change more or less smoothly with the learning rate. However, the training loss or the test accuracy of a single training instant is random. Thus, the choice of a learning rate can be seen as pulling an arm from a continuous-arm bandit, whose expected reward (average test accuracy or average negative training loss) is a smooth function of continuous arm indices. This naturally leads to the Lipschitz bandit problem. We tune the learning rate using the Zooming algorithm for exploration-exploitation in Lipschtiz bandits. We empirically validate our approach and compare it against HyperOpt for a variety of architectures and datasets. The empirical results clearly indicate that our method finds a better learning rate with a lesser number of epochs and fewer evaluations. 
\par 
Batched Lipschitz Exploration (BLiE) is a recently proposed method  \cite{feng2023lipschitz}, which, among all the previous methods, is closest to our approach. It uses a pure exploration bandit algorithm and obtains the learning rate by adaptively partitioning the search space. As discussed later, our approach has much better performance when the computational resource for training, i.e., the number of epochs  and the number of evaluations, are limited, which is often the case in applications like IoT and wireless communication \cite{mao2018deep}.
\section{Proposed Method}
We represent the learning rate as arms of continuous bandits in a known metric space (X, D), where $X = [0,1]$, and D is the metric on X.
At round $t$, a learning rate $x_t$ is sampled and a random reward with mean $\mu(x)$ is observed. The reward is the training loss or test accuracy of the neural network under that learning rate. As was observed in prior work, the mean reward (training loss or accuracy) is a Lipschitz function, i.e.,
\begin{equation}\label{eq1}
    |\mu(x) - \mu(y)|  \leq  D(x,y) 
\end{equation}
for two different learning rates $x$ and $y$. Here $D$ is a metric on $[0,1]$ such that the mean reward is $1$-Lipschitz with respect to $D$.
\par For addressing the Lipschitz multi-arm bandit, there are two primary approaches: static discretization and adaptive discretization of the arms set. 
In the first approach, the space of parameters is uniformly discretized, and a discrete arm multi-armed bandit algorithm is applied on the discretized space. In adaptive discretization, as the name suggests, the discretization is adapted with time based on past samples, the discretization is refined with time for better differentiation between the remaining arms. 
Though the analytical regret guarantees for these two methods are similar in the worst case, the adaptive approach is known to perform better in a typical setting. Hence, in this paper, we adopt the adaptive discretization algorithm, called the zooming algorithm (Algorithm \ref{algzoomingalg})  for optimizing the learning rate.

\par  In the Zooming algorithm, let $n_t(x)$ be the number of times the arm $x$ has been chosen before round t. The confidence radius of arm $x$ at round $t$ is defined as 
\begin{equation} \label{eq2}
   r_t(x) = \sqrt{\frac{2}{n_t(x)+1}} 
\end{equation}

The confidence ball of arm $x$ is a closed ball in the metric space with center $x$ and radius $r_t(x)$.
\begin{equation}\label{eq3}
    B_t(x) = \{ y \in X : D(x, y) \leq r_t(x)\}
\end{equation}
The Zooming algorithm maintains a set $S \subset X$ of “active arms”. 
Active arms at time $t$ are the union of all the balls, $B_t(x)$ for $x \in S$. At time $t$, an arm that is not covered by these balls is included in $S$, assigned an initial confidence radius $1$ and accordingly a  confidence ball.
An active arm with the largest index is selected for playing,  where index at time $t$ for an active arm $x$ is defined as

\begin{equation} \label{eq4}
    index_t(x) = \mu_t(x) + 2r_t(x)
\end{equation}
 \par As mentioned before, we use the zooming algorithm for optimizing the learning rate. We treat learning rates as arms of a continuous bandit on $[0,1]$ and the mean square training error or test accuracy as a reward.

\begin{algorithm}[t]
\caption{\textbf{Zooming algorithm}}
\label{algzoomingalg}
\begin{algorithmic}
        \State \text{\textbf{Initialize:} Set of active arms S $\leftarrow\varnothing$ }
        \For {\text{\textit{each round} t = 1,2,3,....}}
        \State \textit{Activation rule:} Identify arms not covered by $B_t(x)$ for
        \State{ some $x \in S$: activate any such arm, i.e., include in $S$.}
        \State \textit{Selection rule:} Play an active arm with the largest 
        \State{index, defined in \eqref{eq4}.}
        \EndFor
\end{algorithmic}
\end{algorithm}

\section{{Simulation results}}

Our main goal is to test whether the Zooming algorithm is able to find a better learning rate when compared with popular methods such as HyperOpt \cite{bergstra2015hyperopt}. 
We use two different metrics to compare the optimization strategies, i.e., best-found value and best trace. In best trace, while comparing the algorithms, we choose the learning rate with the least area under the convergence curve (AUC). In best found value, the training loss achieved is used as the metric for comparison.
\par So, to compare the HyperOpt with the Zooming algorithm, we ran different experiments with the same number of evaluations for both HyperOpt and the Zooming algorithm for single and multi-hidden layer neural networks for various datasets.

\subsection{Single hidden layer neural network on simulated dataset}
 We start our comparison with a simple multivariate normal dataset on the Teacher-student network which is similar to that of the \cite{Tholeti_2020}. 
 \par In the teacher-student network, there is an underlying network known as the teacher network with weights $W^*$. The weights of the teacher network are sampled from a zero mean unit variance Gaussian distribution i.e., $W^* (i) \sim N(0, 1)$.
 Random samples drawn from multivariate normal distribution where, $X(i) \sim N(0, I)$, for $i = 1,2, \ldots N$, are input to the teacher network. The corresponding output data $Y(i)$ are recorded.  
 \par Next, another neural network with single hidden layer, known as the student network, is trained on this dataset. The weights of the student network are initialized with samples drawn from a zero mean unit variance Gaussian distribution i.e., $W (i) \sim N(0, 1)$. For training the student network, mean square error(MSE) loss function is employed.
 The optimization algorithm used for training is gradient descent (GD).

In the Zooming algorithm, the learning rate samples correspond to arm values, 
and reward values are formulated as negative of MSE.
The neural network is trained with the above-mentioned teacher-student network parameters for each randomly taken arm value and finds the best learning rate using the best trace metric. 
\subsubsection{Comparison of Zooming Algorithm and HyperOpt Algorithm}
A student network with different number of neurons in the input layer ($d$) and in the hidden layer ($k$) is trained for $100$ epochs using GD as the optimization algorithm with N=10,000 data points. 
HyperOpt and Zooming algorithms are compared using the best trace metric. 
 We present results for a specific configuration with ReLU and sigmoid activation functions. Both HyperOpt and Zooming are constrained by a budget on the number of evaluations. 

\subsubsection{Observations for Teacher-Student network}
\begin{figure}[!t]
    \centering 
\includegraphics[width=\columnwidth]{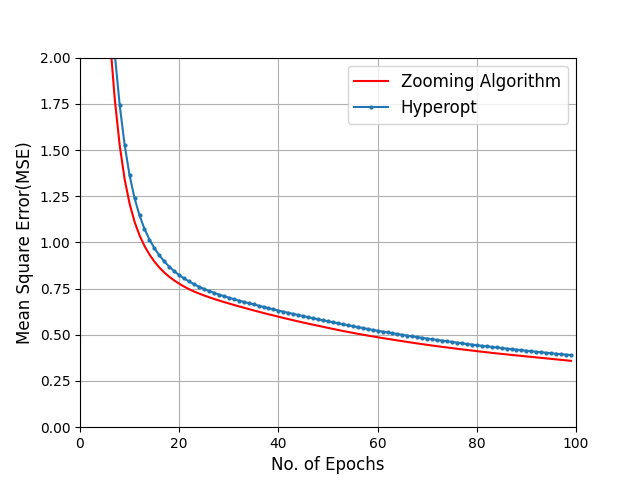}
    \caption{ Best trace comparison of Zooming algorithm and HyperOpt for Single hidden layer ReLU network}
    \label{ReLUloss}
\end{figure}

\begin{figure}[!t]
    \centering 
    \includegraphics[width=\columnwidth]{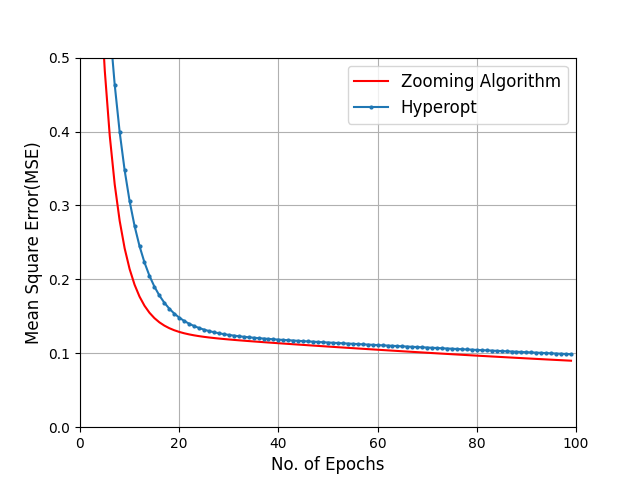}
    \caption{ Best trace comparison of Zooming algorithm and HyperOpt for Single hidden layer Sigmoid network}
    \label{sigmoidloss}
\end{figure}

For each method (Zooming and HyperOpt), we performed $50$ experiments and chose the MSE vs number of epcohs curve with the best trace. These curves are compared  in Fig. \ref{ReLUloss} and Fig. \ref{sigmoidloss}, for ReLU and sigmoid activation, respectively. 
It is seen from the figures that the Zooming algorithm results in better convergence of MSE.
\par We also observed that Zooming is faster in terms of the number of evaluations needed for selecting the best learning rate. Zooming chooses the best learning rate within three evaluations, whereas HyperOpt took ten evaluations. Thus, in the teacher-student network, Zooming has lower training cost and faster convergence than HyperOpt.
 We observed that, in case of ReLU activation, in some cases GD diverges even for the best learning rate chosen by HyperOpt and Zooming. However, such behavior  is not observed for sigmoid activation. The fraction of times HyperOpt and the Zooming algorithm diverge for ReLU activation is given in Table~\ref{tablerelu}. It can be seen that Zooming algorithm has a smaller fraction of divergence than HyperOpt and does not diverge for two architectures: $d=k=10$ and $d=20,k=5$. By further modifying the confidence radius of an arm value in the Zooming algorithm from $r_t(x)$ to $\frac{r_t(x)}{10}$, we observed that the GD in Zooming does not diverge anymore for ReLU. Intuitively, a smaller confidence radius allows for better exploration and thus, Zooming is able to find a learning rate closer to the optimal. We carry this intuition to experiments on other datasets and use this modified version of the Zooming algorithm.


\begin{table}[!t]
\centering
\caption{  Fraction of times Zooming and Hyperopt algorithm diverges for single hidden layer network with ReLU activation }
\begin{tabular}[width=\columnwidth]{|c|c|c|c|c|}
\hline
& & &\multicolumn{2}{c|}{\textbf{No of evaluations}}\\
\hline
\textbf{\textit{Algorithm}} &\textbf{\textit{ d}}  & \textbf{\textit{k}} &\textbf{5}& \textbf{10}  \\
\hline
\multirow{3}{*}{HyperOpt} 
   & 10  &10 &0.37 & 0.38 \\ 
&  20 & 5 & 0 & 0 \\ 
& 5& 20 &0.11&0.02 \\
\cline{1-5} 
\multirow{3}{*}{Zooming}  & 10  &10 &0 & 0 \\
&20&5 & 0 & 0 \\
&5& 20 &0.4&0.22 \\
\cline{1-5} 
\multirow{3}{*}{Zooming with modified radius of arm}  & 10  &10 &0 & 0 \\
&  20 & 5 & 0 & 0 \\
&5& 20 &0&0 \\
\hline
\end{tabular}
\label{tablerelu}
\end{table}


\subsection{Multi-layer feed forward neural network on CIFAR-10 dataset}
HyperOpt and Zooming are used for training multi-layer neural networks of different depths on CIFAR10 dataset and their performance are compared using the best trace metric. 
The number of neurons in the input layer is adjusted according to the dataset and the number of neurons in the output layer is 10. The number of hidden layers is varied between $1$, $3$ and $5$ and the number of neurons in each hidden layer is 50. All activation functions are ReLU. Since we are interested in the regime where the computational budget is limited, we consider the number of evaluations to be $5$ and $10$. Results are tabulated in Table~\ref{comparecifar TABLE}.
The learning rate with the lowest area under the learning curve (AUC) is the best method. It is apparent that Zooming is better than HyperOpt.

\begin{table}[!b]
\centering
\caption{ Comparision of Zooming and HyperOpt algorithms for CIFAR10 dataset with ReLU activation}
\resizebox{\columnwidth}{!}{
\begin{tabular}[width=\columnwidth]{|c|c|c|c|c|c|}
\hline
\textbf{{layers}} & \textbf{{ Evals }} & \textbf{{ Algorithm }}  & \textbf{Learning rate} & \textbf{AUC} & \textbf{Best method}   \\
\hline
\multirow{4}{*}{1} &
  \multirow{2}{*}{5} & Zooming & 0.02419 & 68.4664 &  \multirow{2}{*}{\text{\textbf{Zooming}}} \\
& &  HyperOpt & 0.08626 &  69.7590 &    \\
\cline{2-6} 
&  \multirow{2}{*}{10} & Zooming &0.04695  & 68.2776 &  \multirow{2}{*}{\text{\textbf{Zooming}}}\\
&  & HyperOpt  & 0.1176 & 72.2055 & \\
\hline
\multirow{4}{*}{3} &
  \multirow{2}{*}{5} &  Zooming &0.11863 & 58.8852 &  \multirow{2}{*}{\text{\textbf{Zooming}}}\\
& &  HyperOpt & 0.02939 & 62.3642 &    \\
\cline{2-6} 
&  \multirow{2}{*}{10} & Zooming & 0.0879  & 69.4581  &  \multirow{2}{*}{\text{\textbf{Zooming}}} \\
&  & HyperOpt  & 0.15002  & 77.5124& \\
\hline
\multirow{4}{*}{5} &
  \multirow{2}{*}{5} &  Zooming & 0.11084  & 63.6210  &  \multirow{2}{*}{\text{\textbf{Zooming}}}\\
& &  HyperOpt & 0.13833  & 64.0647&    \\
\cline{2-6} 
 &
  \multirow{2}{*}{10} & Zooming & 0.06707  & 65.3482  &  \multirow{2}{*}{\text{\textbf{Zooming}}} \\
& & HyperOpt  & 0.06695 & 65.5647& \\
\hline
\end{tabular}
}
\label{comparecifar TABLE}
\end{table}
\subsection{Comparison of Zooming and HyperOpt for ResNet20 Architecture}
To test the Zooming algorithm on deeper architectures, we first choose  ResNet20 architecture with same network parameters as in \cite{he2016deep}.
ResNet20 is trained for 200 epochs on CIFAR10 and CIFAR100 datasets, with a batch size of 128. Stochastic gradient descent (SGD) with momentum is used as the optimization algorithm along with a learning rate scheduler which reduces the learning rate at epochs $80$, $120$ and $160$. 

\par HyperOpt and Zooming  are compared for ResNet20 architecture on the CIFAR-10 and CIFAR100 datasets and the results are tabulated in Table \ref{resnet20}. Since we are interested in training under computation and memory budget, we allow at most twenty evaluations. We observed that learning was not up to the mark below that budget.

\subsubsection{Observations for CIFAR-10 dataset with ResNet-20 architecture} 
\begin{figure}[!t]
    \centering 
   \includegraphics[width=\columnwidth]{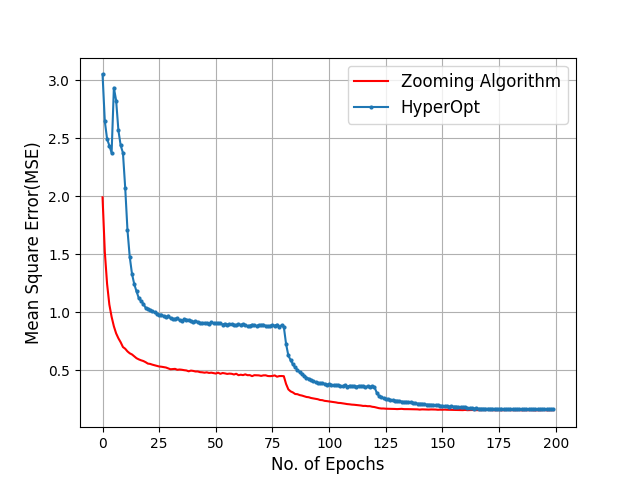}   
    \caption{ Best trace comparison of Zooming algorithm and HyperOpt for ResNet20 network architecture with CIFAR10 dataset.}
    \label{resnet FIG}
\end{figure}

The best trace curve for $20$ evaluations is plotted in Fig.~\ref{resnet FIG}  for Zooming and HyperOpt. It is observed that Zooming has faster convergence than HyperOpt. 

In Table \ref{resnet20}, different performance metrics of Zooming and HyperOpt for ResNet20 on CIFAR10 and CIFAR100 are compared. The third column gives information about the number of learning rate samples chosen by Zooming and HyperOpt algorithms for finding the best learning rate. AUC are presented in the fourth column, while the test accuracy of the learned models are reported in the fifth column. We observe that Zooming performs better in all the metrics. Furthermore,  Zooming is able to identify a good learning rate using  three learning rate samples whereas HyperOpt needed twenty samples. This implies that under Zooming the training cost is also lower. 

\begin{table}[!t]
\centering
\caption{Comparision of Zooming and HyperOpt algorithm for ResNet20 architecture}
\resizebox{\columnwidth}{!}{
\begin{tabular}{|c|c|c|c|c|c|}
\hline
\multicolumn{6}{|c|}{\textbf{ResNet20 architecture results for CIFAR10 dataset} }\\
\hline
\textbf{\textit{Dataset}}&\textbf{\textit{Algorithm}}& \textbf{\textit{No. of}} &  \textbf{\textit{AUC}}&\textbf{\textit{Test}}&\textbf{\textit{Best}}\\
&&\textbf{\textit{samples}}&&\textbf{\textit{Accuracy}}&\textbf{\textit{Algorithm}} \\
\hline

\multirow{2}{*}{CIFAR10}& Zooming &3& 67.136 & \textbf{91.92} & \multirow{2}{*}{\textbf{Zooming}}\\
& HyperOpt&20&125.262&90.29&\\
\hline
\multirow{2}{*}{CIFAR100}& Zooming &3&242.78 & \textbf{65.10}&\multirow{2}{*}{\textbf{Zooming}}\\
& HyperOpt&20&293.81&62.76&\\
\hline
\end{tabular}
}
\label{resnet20} 
\end{table}

\subsection{Comparision of Zooming with BLiE algorithm for Resnet20 and ResNet56 Architetures}
 Zooming is compared with BliE for different ResNet architectures for CIFAR10 and CIFAR100 datasets, and the results are tabulated in Table \ref{cube}. Again, motivated by training with computation and memory limitation, we restrict to twenty evaluations.

\subsubsection{Observations}
 
We observed that when the number of epochs is  $300$, Zooming is able to reach a test accuracy of 90.34 $\% $ with only $3$ samples, whereas  the BLiE achieves  73.30 $\% $ and needs $5$ samples. For higher number of epochs, on CIFAR10, the test accuracy of both the algorithms are similar, but Zooming achieves that using almost half the samples. On CIFAR100, even at $600$ epochs Zooming is significantly better than BLiE both in terms of test accuracy and number of samples. Thus, under Zooming when the budget for the number of evaluations is low ($20$), test accuracy is better than BLiE while  the training cost is  smaller. We also observed that when the number of epochs is less than $300$, BLiE is unable to find a learning rate that leads to good test accuracy. 

The main reason behind better performance of Zooming when evaluations and epochs are limited  is that it was designed for minimizing regret for exploration-exploitation Lipschitz bandit over relatively fewer trials. On the other hand, BLiE is based on an elimination based exploration algorithm which reaches arbitrarily close to the optimum when  the number of trials is large. However, this exploration algorithm is not necessarily efficient when the number of trials is limited. 

\begin{table}[!t]
\centering
\caption{Comparision of Zooming and BLiE algorithm for ResNet architectures}
\resizebox{\columnwidth}{!}{
\begin{tabular}{|c|c|c|c|c|c|}
\hline
\multicolumn{6}{|c|}{\textbf{ResNet architecture results for CIFAR10 and CIFAR100 datasets} }\\
\hline
\textbf{\textit{Architecture}}&\textbf{\textit{Total}}& \textbf{\textit{Algorithm}}&\textbf{\textit{No. of}}&\textbf{\textit{Test}}&\textbf{\textit{Best}}\\
\textbf{\textit{\& Dataset}}&\textbf{\textit{Epochs}}& &\textbf{\textit{samples}}&\textbf{\textit{Accuracy}} & \textbf{\textit{Algorithm}} \\
\hline
ResNet20&\multirow{2}{*}{300}&{Zooming} & 3 & \textbf{90.34} &  \multirow{2}{*}{\text{\textbf{Zooming}}}\\
\& CIFAR10&&BLiE& 5 &73.30 &\\
\hline
ResNet20&\multirow{2}{*}{600}& Zooming &3 &\textbf{91.92} &  \multirow{2}{*}{\text{\textbf{Zooming}}}\\
\& CIFAR10&&BLiE& 7 & 91.40 &\\
\hline
ResNet56& \multirow{2}{*}{400} & Zooming & 2  &\textbf{91.92} &  \multirow{2}{*}{\text{\textbf{Zooming}}}\\
\& CIFAR10&&BLiE& 5 & 91.40 &\\
\hline
ResNet56&\multirow{2}{*}{600}& Zooming& 3 & \textbf{68.82} &   \multirow{2}{*}{\text{\textbf{Zooming}}}\\
\& CIFAR100&&BLiE& 7 & 64.52 &\\
\hline
\end{tabular}
}
\label{cube} 
\end{table}

\section{Conclusion}
In this work, 
we propose a method for finding the optimal learning rate  using the Zooming algorithm for Lipschitz bandits.  We show through extensive experiments on simulated data for teacher-student network as well as on CIFAR10 and CIFAR100 datasets for ResNet20 and ResNet56, that the proposed method outperforms HyperOpt and BLiE both in terms AUC and test accuracy. 
Typically, tuning the learning rate, especially for deeper architectures, requires both a) multiple evaluations and b) large number of epochs. Our proposed algorithm is able to find a better learning rate with a lesser number of evaluations and fewer epochs when compared with both HyperOpt and BLiE.
This enables us to perform efficient neural network training with lower computing power and time.

\bibliography{sample_ieee}

\begin{thebibliography}{10}

\bibitem{liu2017survey}
W.~Liu, Z.~Wang, X.~Liu, N.~Zeng, Y.~Liu, and F.~E. Alsaadi, ``A survey of deep neural network architectures and their applications,'' {\em Neurocomputing}, vol.~234, pp.~11--26, 2017.

\bibitem{mao2018deep}
Q.~Mao, F.~Hu, and Q.~Hao, ``Deep learning for intelligent wireless networks: A comprehensive survey,'' {\em IEEE Communications Surveys \& Tutorials}, vol.~20, no.~4, pp.~2595--2621, 2018.

\bibitem{snoek2012practical}
J.~Snoek, H.~Larochelle, and R.~P. Adams, ``Practical bayesian optimization of machine learning algorithms,'' {\em Advances in neural information processing systems}, vol.~25, 2012.

\bibitem{rasmussen2003gaussian}
C.~E. Rasmussen, ``Gaussian processes in machine learning,'' in {\em Summer school on machine learning}, pp.~63--71, Springer, 2003.

\bibitem{hutter2011sequential}
F.~Hutter, H.~H. Hoos, and K.~Leyton-Brown, ``Sequential model-based optimization for general algorithm configuration,'' in {\em Learning and Intelligent Optimization: 5th International Conference, LION 5, Rome, Italy, January 17-21, 2011. Selected Papers 5}, pp.~507--523, Springer, 2011.

\bibitem{bergstra2011algorithms}
J.~Bergstra, R.~Bardenet, Y.~Bengio, and B.~K{\'e}gl, ``Algorithms for hyper-parameter optimization,'' {\em Advances in neural information processing systems}, vol.~24, 2011.

\bibitem{bergstra2015hyperopt}
J.~Bergstra, B.~Komer, C.~Eliasmith, D.~Yamins, and D.~D. Cox, ``Hyperopt: a python library for model selection and hyperparameter optimization,'' {\em Computational Science \& Discovery}, vol.~8, no.~1, p.~014008, 2015.

\bibitem{jamieson2016non}
K.~Jamieson and A.~Talwalkar, ``Non-stochastic best arm identification and hyperparameter optimization,'' in {\em Artificial intelligence and statistics}, pp.~240--248, PMLR, 2016.

\bibitem{li2018hyperband}
L.~Li, K.~Jamieson, G.~DeSalvo, A.~Rostamizadeh, and A.~Talwalkar, ``Hyperband: A novel bandit-based approach to hyperparameter optimization,'' {\em Journal of Machine Learning Research}, vol.~18, no.~185, pp.~1--52, 2018.

\bibitem{falkner2018bohb}
S.~Falkner, A.~Klein, and F.~Hutter, ``Bohb: Robust and efficient hyperparameter optimization at scale,'' in {\em International conference on machine learning}, pp.~1437--1446, PMLR, 2018.

\bibitem{senior2013empirical}
A.~Senior, G.~Heigold, M.~Ranzato, and K.~Yang, ``An empirical study of learning rates in deep neural networks for speech recognition,'' in {\em 2013 IEEE international conference on acoustics, speech and signal processing}, pp.~6724--6728, IEEE, 2013.

\bibitem{nesterov2004introductory}
Y.~Nesterov, ``Introductory lectures on convex programming: a basic course, volume i,'' 2004.

\bibitem{Tholeti_2020}
T.~Tholeti and S.~Kalyani, ``Tune smarter not harder: A principled approach to tuning learning rates for shallow nets,'' {\em IEEE Transactions on Signal Processing}, vol.~68, p.~5063–5078, 2020.

\bibitem{feng2023lipschitz}
Y.~Feng, W.~Luo, Y.~Huang, and T.~Wang, ``A lipschitz bandits approach for continuous hyperparameter optimization,'' {\em arXiv preprint arXiv:2302.01539}, 2023.

\bibitem{he2016deep}
K.~He, X.~Zhang, S.~Ren, and J.~Sun, ``Deep residual learning for image recognition,'' in {\em Proceedings of the IEEE conference on computer vision and pattern recognition}, pp.~770--778, 2016.

\end{thebibliography}
\bibliographystyle{ieeetr}

\end{document}